\documentclass[lettersize,journal]{IEEEtran}
\usepackage{amsmath,amsfonts}
\usepackage{algorithmic}
\usepackage{algorithm}
\usepackage{array}
\usepackage{subcaption} 
\usepackage{textcomp}
\usepackage{stfloats}
\usepackage{url}
\usepackage{verbatim}
\usepackage{graphicx}
\usepackage{cite}
\usepackage{xcolor} 

\usepackage{float}

\begin{document}

\title{\textsc{Stereofog} - Computational DeFogging via Image-to-Image Translation on a real-world Dataset}

\author{Anton Pollak, Rajesh Menon,~\IEEEmembership{Member,~IEEE, Senior Member,~IEEE}}

\maketitle

\begin{figure*}[b]
\centering\includegraphics[width=0.9\textwidth]{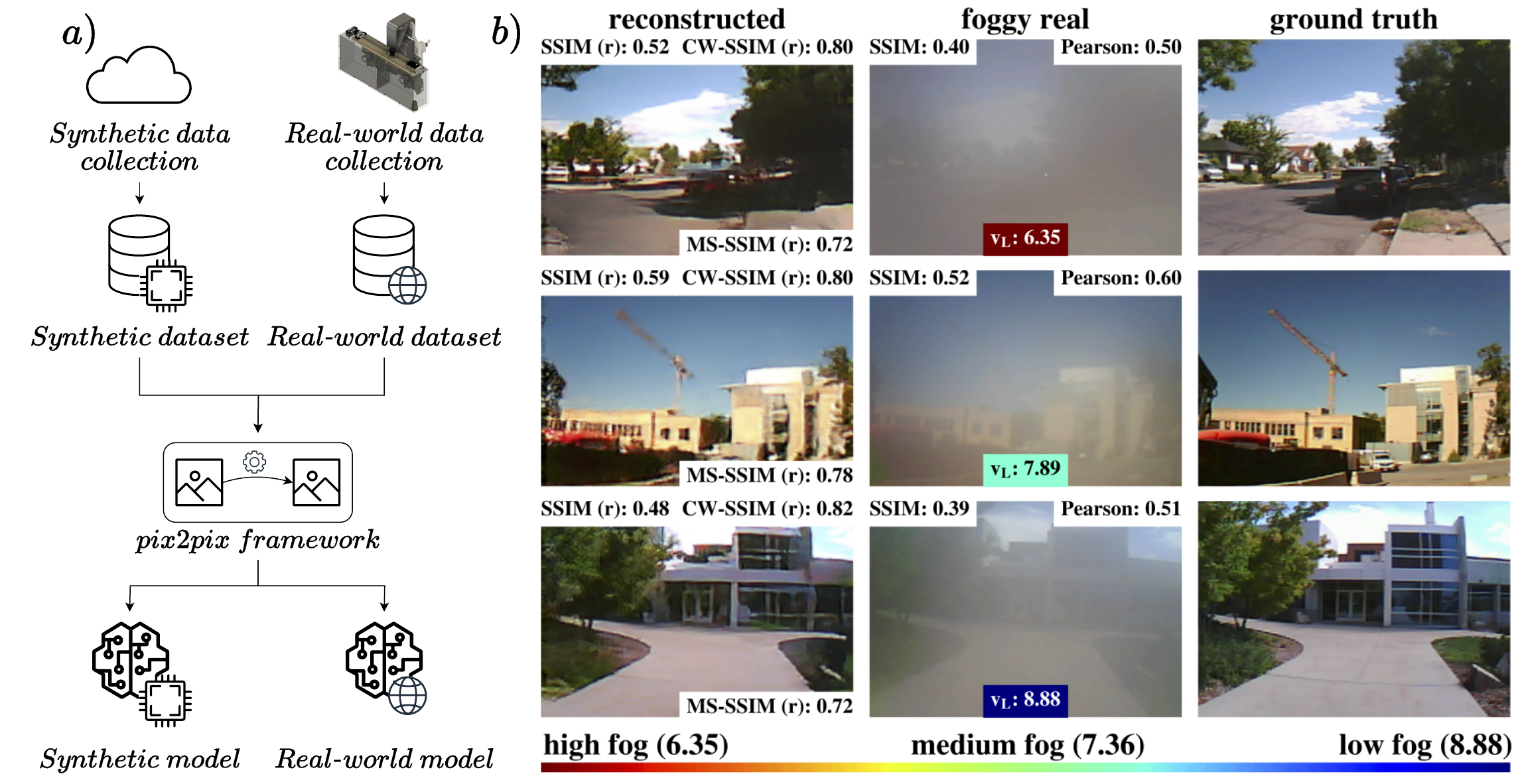}
\caption{Overview of the \textsc{Stereofog} project. a): A diagram summarizing the work done in this work. b): Example results obtained by applying the \texttt{pix2pix} framework to the \textsc{Stereofog} dataset. Our approach works for a range of fog densities.}
\label{fig:headline}
\end{figure*}

\begin{abstract}
Image-to-Image translation (I2I) is a subtype of Machine Learning (ML) that has tremendous potential in applications where two domains of images and the need for translation between the two exist, such as the removal of fog. For example, this could be useful for autonomous vehicles, which currently struggle with adverse weather conditions like fog. However, datasets for I2I tasks are not abundant and typically hard to acquire. Here, we introduce \textsc{Stereofog}, a dataset comprised of $\mathbf{10,067}$ paired fogged and clear images, captured using a custom-built device, with the purpose of exploring I2I's potential in this domain. It is the only real-world dataset of this kind to the best of our knowledge. Furthermore, we apply and optimize the \emph{pix2pix} I2I ML framework to this dataset. With the final model achieving an average Complex Wavelet-Structural Similarity (CW-SSIM) score of $\mathbf{0.76}$, we prove the technique's suitability for the problem. 
\end{abstract}

\begin{IEEEkeywords}
Image defogging, Image-to-image translation, pix2pix, image dataset.
\end{IEEEkeywords}

\vfill\eject
\section{Introduction}
\IEEEPARstart{L}{ow} visibility caused by fog accounts for over 38,700 vehicle crashes on US roads every year that result in 600 annual deaths, according to the Federal Highway Administration \cite{FederalHighwayAdministration2023}. Meanwhile, the transition to self-driving cars is well underway (see Fig. S1 in supplement). Within the 5 levels of automation defined by the Society of Automotive Engineers (SAE), level 4 autonomous driving is the first level where the driver does not have to be engaged with driving while the system is in operation \cite{NationalHighwayTrafficSafetyAdministration}. Even in the conservative estimate, level 4 and 5 are estimated to make up $8\%$ of the US vehicles sales in 2035.

However, just as humans, current autonomous vehicles struggle with adverse weather conditions. In fact, this problem is now recognized as the major barrier for achieving level 4 autonomy \cite{Zhang2023,Zang2019}. Being able to see through fog clearly would significantly mitigate this problem. It would be useful in many other areas as well, such as in search and rescue missions, particularly involving autonomous agents.

Computational defogging via Machine Learning (ML) could be a powerful solution. Image-to-Image translation (I2I) is particularly well-suited for the task of translating the fogged image to the clear image. To achieve optimum results, I2I should be applied to a paired-image dataset, meaning that each clear image corresponds to exactly one fogged image (more information on this in section \ref{sec:sota-I2I}). However, such a dataset is much harder to acquire than an unpaired one. The reason for this is that the images have to be exactly paired. This practically rules out already existing datasets that were not collected with this use case in mind. The defogging task is especially challenging, since with naturally occurring fog, capturing the same scene with and without fog is generally not possible. There are synthetic datasets with computationally generated fog. However, it is not clear whether these are sufficiently good substitutes for the real-world fog behavior, which can involve very complex hydrodynamics.

Here, we apply an I2I ML model to computationally defog real-world images. In order to acquire the requisite paired-image dataset, we created a custom device that consists of two cloned cameras imaging the same scene and an enclosure that introduces fog in front of only one. Thereby, we created a dataset containing $10,067$ image pairs. It is the only real-world dataset of this kind to the best of our knowledge. Then, we trained a \verb|pix2pix| I2I model to perform the translation \emph{fog} $\rightarrow$ \emph{no fog}. Nine of the model's hyperparameters were evaluated to find the best-performing configuration.
The full dataset, Supplement 1 as well as accompanying code can be found on this project's Github page \cite{Pollak2023} (\url{https://github.com/apoll2000/stereofog}).

\IEEEpubidadjcol

\section{Methods}
\subsection{State of the Art \& Technology}

\subsubsection{Fog datasets}
\label{sota:fog-datasets}
Datasets with foggy images exist, however none of them serve our purpose of computational defogging effectively. For example, \textsc{Bijelic et al.} collected 12.000 images of real-world driving in adverse weather for the "Seeing Through Fog" project, carried out as part of the DENSE project of the European Union \cite{Bijelic2020}. In addition to RGB images, they also recorded thermal and LIDAR data. The \emph{Foggy Driving} dataset by \textsc{Sakaridis et al.} contains 101 images of foggy scenes \cite{Sakaridis2018}. However, in all cases, no paired-clear images are available for supervised learning.

There are synthetic datasets that insert fog into images computationally, typically using depth information about a scene. In order to attain paired clear-foggy images for semantic segmentation, \textsc{Sakaridis et al.} used the autonomous driving dataset \emph{Cityscapes} that includes precomputed depth maps to computationally inject fog into clear images \cite{Cordts2016}. This dataset named "Foggy Cityscapes" contains 15.000 images \cite{Sakaridis2018}. When no depth information is available, \textsc{Nie et al.} used the self-supervised monocular depth prediction model \emph{Monodepth2} in order to achieve an estimated depth map that can be used for fog generation. They used this method to improve the accuracy of lane-detection algorithms \cite{Nie2022}.

In cooperation with researchers from University of Tübingen, Germany, we used their fog simulator implemented in OpenGL to photorealistically render fog and snow into clear images, based on the corresponding depth maps \cite{Bernuth2019}. In addition, they also provided us simulated foggy scenes from the research driving simulator Car Learning to Act (CARLA) \cite{Dosovitskiy2017}. The advantage of this approach is that since the scenes are rendered inside the simulator, it is possible to obtain perfect depth information, which improves the accuracy of the defogging algorithm.

All three synthetic fog datasets used in this project contain images with varying levels of fog density. And in all cases, the efficacy of a defogging model trained on synthetic images, on real-world foggy images is unproven (see Supplement 1 for an example). 

\subsubsection{Haze datasets \& Dehazing algorithms}
A similar phenomonen to fog is haze. Several datasets for image dehazing exist. Popular examples include the RESIDE dataset with its subsets for indoors and outdoors \cite{Li2019}, the Haze4K dataset \cite{Liu2021} or the HazeRD dataset \cite{Zhang2017}. Dehazing algorithms applied to these datasets include MixDehazeNet \cite{Lu2023}, SFNet \cite{Lu2023} and DEA-Net \cite{Chen2023}. The models usually work with deep neural networks and are attention-based. However, only synthetically generated haze and low haze thicknesses are used (see Figure S3 in Supplement 1). Once again, the real-world efficacy of models trained on synthetic haze is unproven.

\subsubsection{Image-to-image translation}
\label{sec:sota-I2I}
Image-to-Image translation (I2I) is "the process of transforming an image from one domain to another, where the goal is to learn the corresponding mapping" \cite{Suarez2021}. In our case, the domains correspond to "foggy" and "clear" images. Convolutional neural networks (CNNs) are commonly used in many image processing tasks. However, CNNs require a good loss function, which may not be readily possible\cite{Isola2018}. Instead, Variational Autoencoders (VAEs) can be used \cite{Pang2021}. Another popular architecture is the Generative Adversarial Network (GAN),\cite{Goodfellow2014} which consists of a generator and a discriminator. The discriminator attempts to identify fake images, which are created by the generator, and thereby implicitly learns the loss function. GANs, however, offer limited control over their output, since the input to the generator is a random noise vector. To mitigate this issue, Conditional GANs were introduced, whose architecture allows for input of additional information to the generator during training \cite{Pang2021}.

I2I can be performed with both paired (supervised) and unpaired (unsupervised) data (models). Unpaired data are much easier to acquire in the real world, an example of which is the photo-to-painting translation \cite{Pang2021}. However, as we show in section 8 of the supplement, models trained on unpaired data perform significantly worse than those trained on paired data. Furthermore, a popular model for the latter task is \verb|CycleGAN|\cite{Zhu2020}, which includes a reverse constraint, requiring that the reverse image, translated from the goal domain back to its original domain, maintains optimal similarity to the original image. Such a network thus implicitly applies physics-based constraints. This approach can be applied to paired-image data as well\cite{Isola2018}, which is what we use in this work. Both frameworks build on the aforementioned Conditional GANs.

\subsubsection{Image comparison metrics}
\label{sec:sota-image-comparison}
When comparing I2I models, it is necessary to evaluate their performance without bias. Therefore, the reconstructed and ground truth images need to be compared using some type of similarity metric. A common method is to compare the two images pixel-by-pixel \cite{Veldhuizen1998}. One measure of this type is the \emph{Mean Square Error (MSE)}. This measure is bit-depth-specific, which makes comparing results for images with different bit depths difficult \cite{Veldhuizen1998}. Therefore, the \emph{Peak Signal-to-Noise Ratio (PSNR)} was introduced as a bit-depth-agnostic measure that scales the MSE to the pixel range \cite{Veldhuizen1998}. Both measures can give misleading results that disagree with common sense (see, for example, \cite{Wang2004} or \cite{Wang2009}). Two other pixel-based metrics are the \textsc{Pearson} correlation coefficient \cite{ChicagoBoothCenterforDecisionResearch} and the \emph{Normalized Cross Correlation Coefficient (NCC)} \cite{Winkelmann2018}. Another metric that focuses on structural information is the \emph{Structural Similarity Index (SSIM)}, introduced in 2004 by \textsc{Wang et al.} The disadvantage of this metric is that it is fairly sensitive to rotations and spatial shifts in the image \cite{MSUGraphics&MediaLabVideoGroup2021}. For our problem, these shifts will always be present (see section \ref{sec:device}), which leads to slight variations in the image perspectives. A measure that is less sensitive to these shifts is the \emph{Complex Wavelet-SSIM (CW-SSIM)}, \cite{Wang2005} which is what we use in this paper. Finally, the \emph{Multi-scale SSIM (MS-SSIM)} \cite{Wang2003} was also implemented for completeness. 

Since the loss function of an I2I model essentially compares images, these metrics can be adapted into loss functions. We implemented this as part of the hyperparameter-tuning step (see section \ref{sec:ml-model-hyperparam}).

\subsection{Binocular camera setup}
\label{sec:device}
\begin{figure}[htbp]
\centering
\begin{subfigure}[b]{\linewidth}
\includegraphics[width=\linewidth]{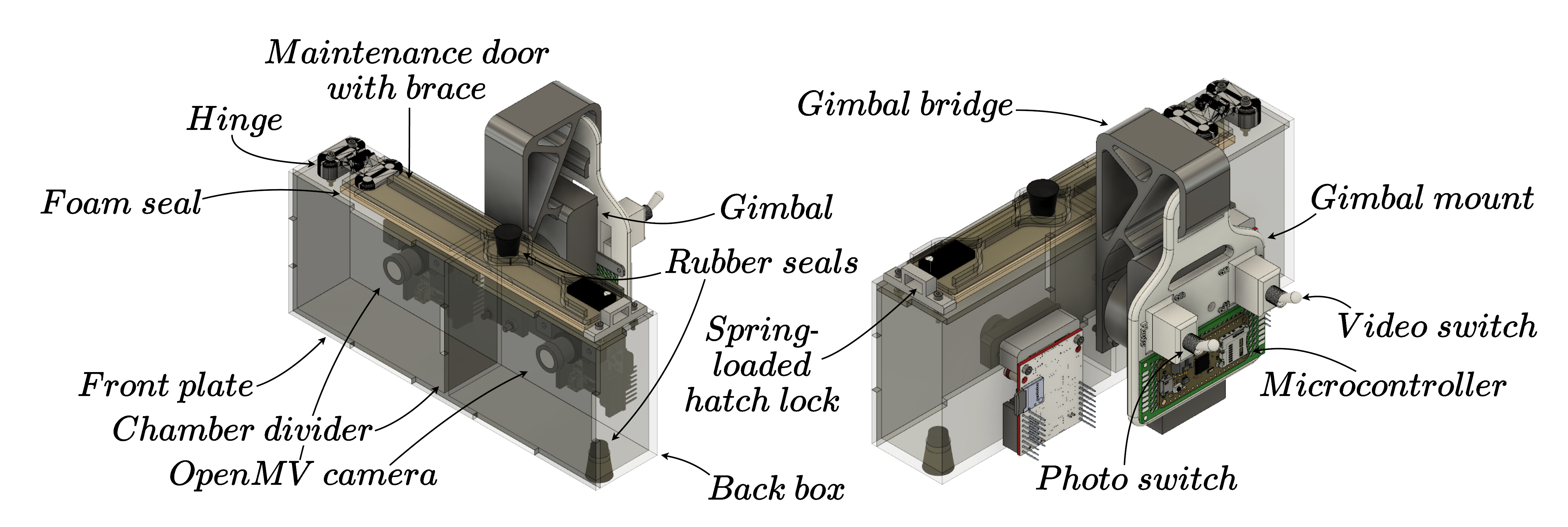}
\end{subfigure}
\begin{subfigure}[b]{.49\linewidth}
\includegraphics[width=\linewidth]{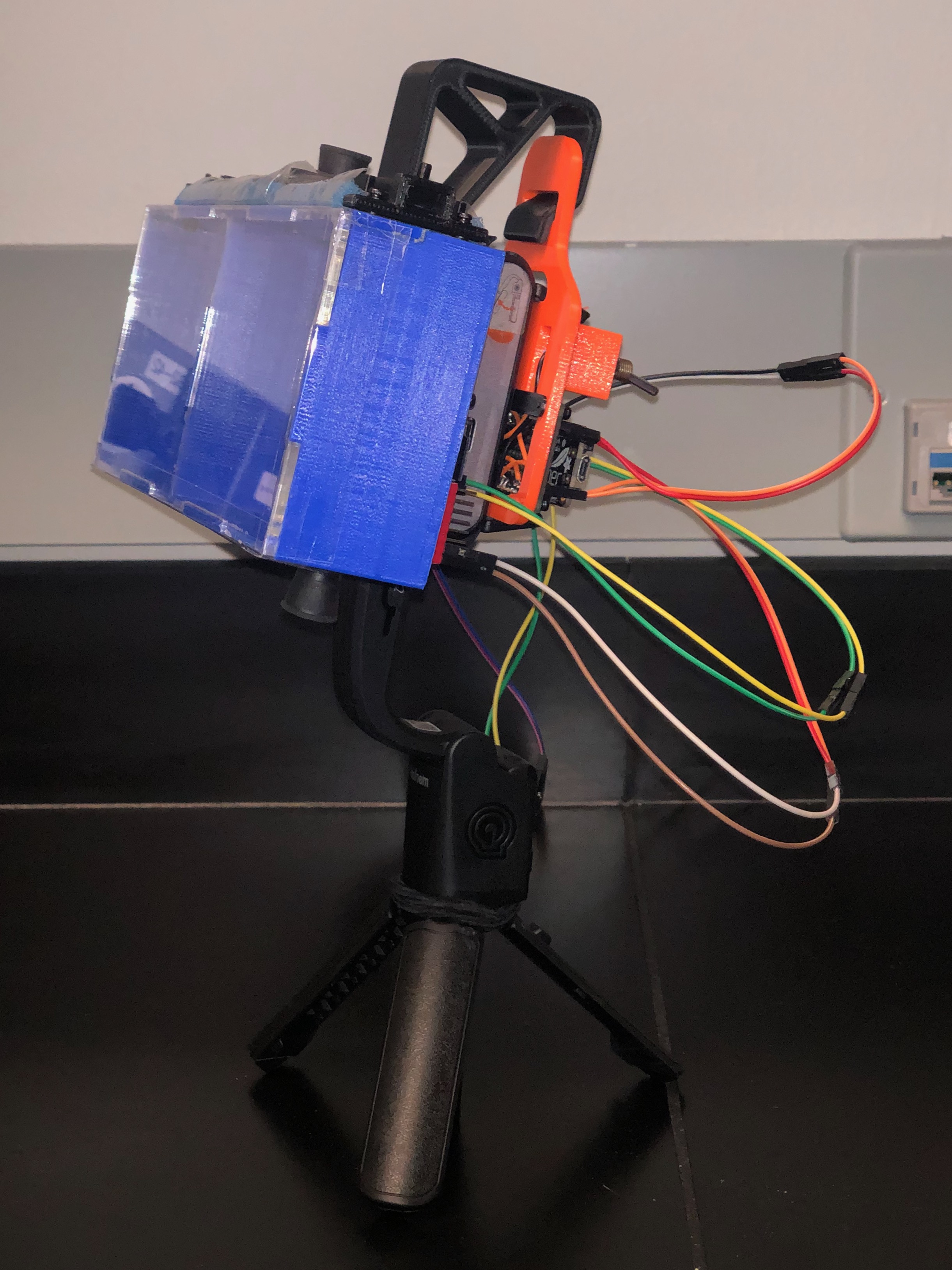}
\end{subfigure}
\begin{subfigure}[b]{.49\linewidth}
\includegraphics[width=\linewidth]{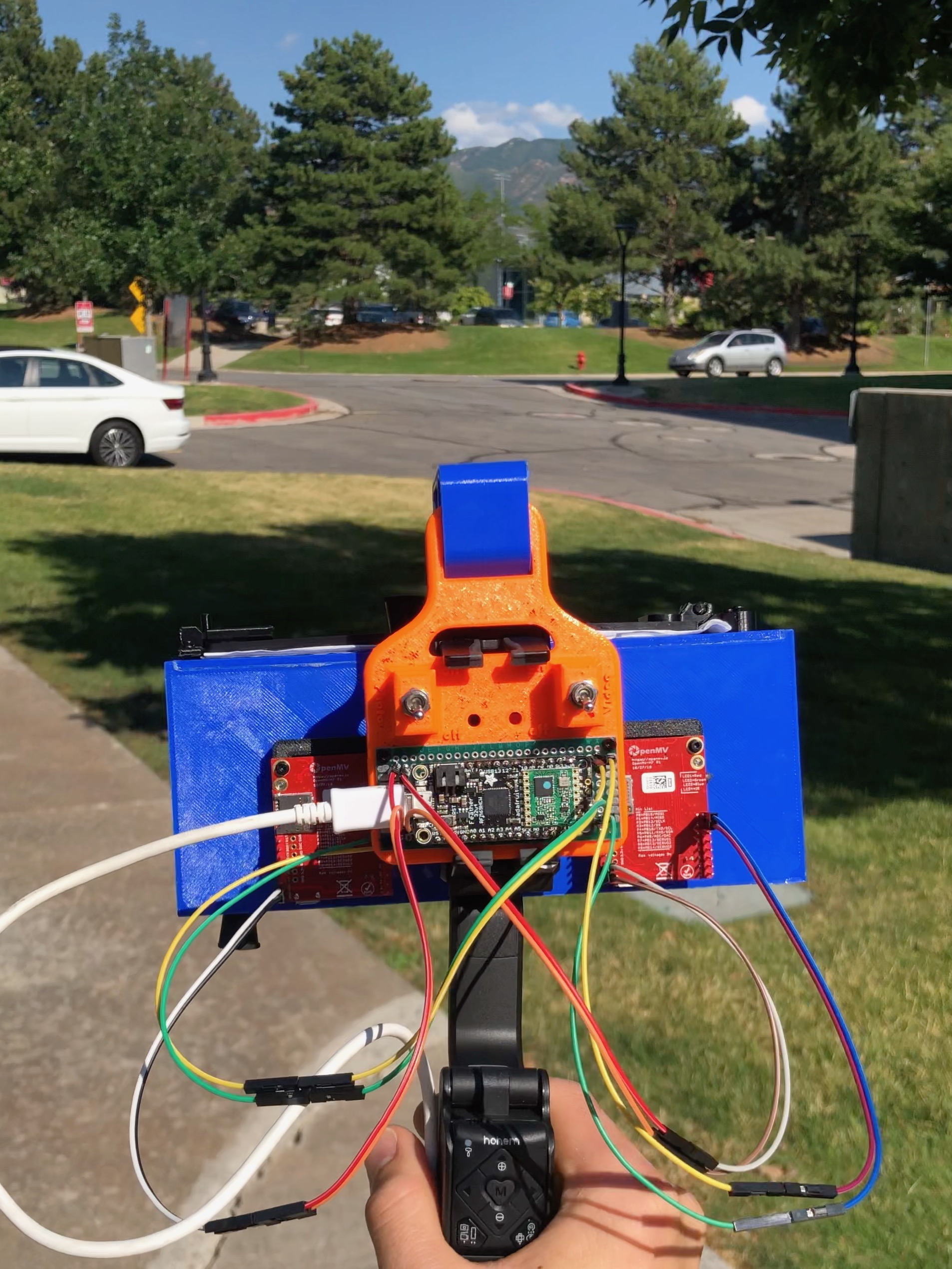}
\end{subfigure}
\caption{Binocular camera setup to capture foggy-clear image pairs. Top: Labelled CAD model. Bottom: Photographs of the setup.}
\label{fig:fog-device-images}
\end{figure}

In order to collect the \textsc{Stereofog} dataset, a binocular camera setup was built (Fig. \ref{fig:fog-device-images}). The setup is comprised of 2 cloned cameras in their separate compartments imaging the same scene. One of the compartments is filled with fog, while the other is left clear. The cloned cameras must be synchronized and the setup needs to be portable for easy data collection. We mounted two \emph{OpenMV H7} cameras \cite{OpenMVLLC2023} onto a custom designed 3D-printed pair of chambers, which was sealed off in the front by a laser cut clear acrylic front plate. On the top, a foam-sealed hinged door was fitted to allow for easy access for cleaning. The fog chamber included two holes to insert fog, and to allow excess air to escape. Both openings are sealable using rubber stoppers.

 The setup was mounted onto a single-axis gimbal for image stabilization. A custom mount was designed for the gimbal to make it compatible with different models. Because the gimbal is only designed for the mechanical load of a smartphone, the setup was connected to the gimbal mount using a 3D printed bridge. This approximately aligned its center of mass with the gimbal's axis of rotation, minimizing the resulting torque on the motor, and thus keeping it within the gimbal's operational limits.

To trigger the two cameras at the same time and record paired images, an \emph{Arduino} microcontroller was used to periodically send a capture signal to the programmable cameras, on which \emph{Python} code evaluated the signal, captured an image and saved it. One toggle switch was used to enable and disable the  phototrigger signal, while another one was used to trigger a video recording on both cameras. Both switches and the microcontroller were housed inside the gimbal mount. Power was supplied using a powerbank, and the microcontroller relayed this voltage to the cameras.

\subsection{Dataset}
\label{sec:methods-dataset}
We collected a total of $10,067$ image pairs in August and September 2023 on the campus of University of Utah in Salt Lake City. Table S1 in Supplement 1 summarizes the different subsets (named "runs") of collected data. The nomenclature followed the collection date and was formatted according to \textsc{ISO 8601}, as well as the run index for each day. The accompanying datasheet \verb|stereofog_dataset_metadata.csv| contains details. Each run contains the subfolders A and B, corresponding to the clear and foggy images, respectively. Each file has a unique name composed of the parent folder's name and its index within the subset (e.g., \verb|2023-08-25_RUN1__114.bmp|).
\begin{figure}[htbp]
\centering
\begin{subfigure}[b]{0.99\linewidth}
\includegraphics[width=\linewidth]{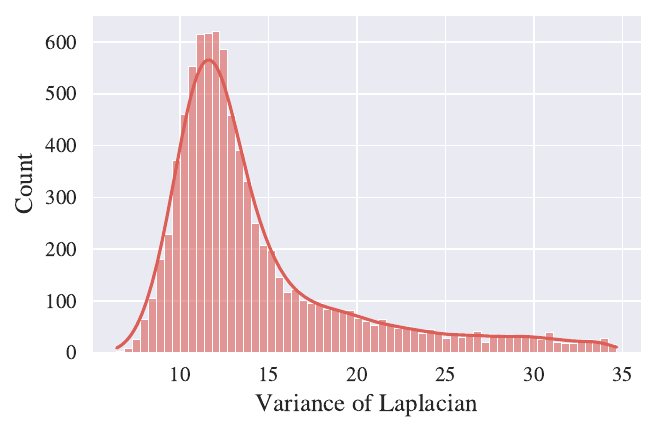}
\end{subfigure}
\begin{subfigure}[b]{\linewidth}
\includegraphics[width=\linewidth]{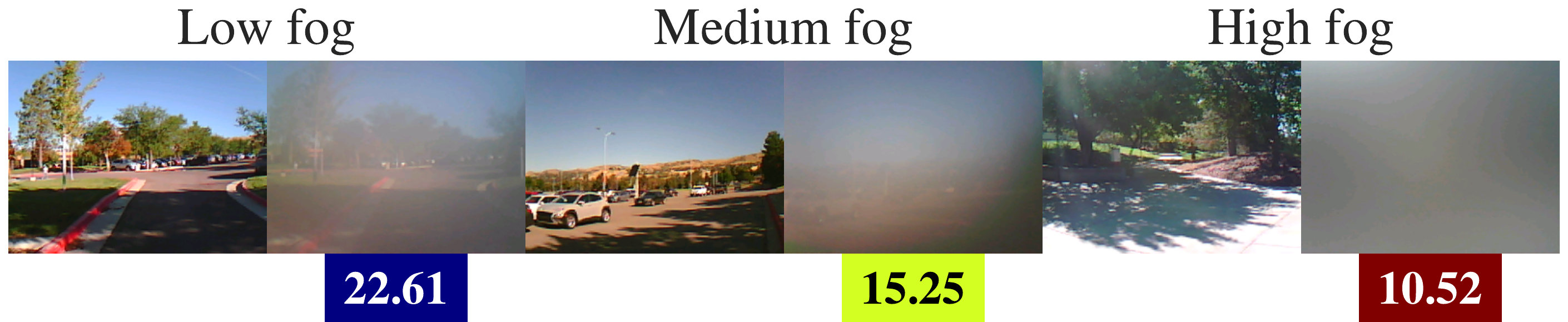}
\end{subfigure}
\caption{Distribution of the Variances of the Laplacian ($\text{v}_\text{L}$) for the \textsc{Stereofog} dataset. Sample image pairs from the dataset with different fogginess levels and their respective $\text{v}_\text{L}$ values for context}
\label{fig:laplacian-stats}
\end{figure}
We used the variance of the Laplacian \cite{Bansal2016, Kinght2018} (labeled as $\text{v}_\text{L}$) to quantify fogginess, since image blurring is a common effect of fog. Figure \ref{fig:laplacian-stats} shows the distribution of the Laplacian values within our dataset (lower values corresponding to denser fog). Outliers have been removed using the interquartile rule in order to make the plot more informative, retaining $85.94\%$ of the data points. The distribution clearly peaks around $12$. The bottom subplots puts these Laplacian values into perspective with examples.

\subsection{Machine Learning}
\subsubsection{Dataset augmentation}
For the training of the final model, the dataset was augmented using the \emph{Augmentor} Python library \cite{Bloice2023}. The augmentation techniques applied were a left-right flip and random zoom with probability of $30\%$. Only the training subset of the data was augmented to prevent data leakage. For the hyperparameter tests, the training was performed on the non-augmented data to keep training times reasonable.

\subsubsection{Model \& Hyperparameters}
\label{sec:ml-model-hyperparam}
The dataset was first preprocessed to be consistent with \verb|pix2pix| (see \ref{sec:sota-I2I}). \verb|pix2pix| possesses various tunable hyperparameters associated with the generator, discriminator, GAN and preprocessing. Table S2 in Supplement 1 lists details for the hyperparameters that were optimized (namely normalization, loss function, netD type, \# of layers in the discriminator, netG type, GAN mode, \# of generative filters in the last convolutional layer, \# of discriminative filters in the first convolutional layer, and network initialization type). All models were undertrained with only 25 epochs and 15 additional decay epochs, in order to reduce training time. 

\subsubsection{Evaluation}
When performing hyperparameter evaluation and having to determine the quality of the model, visual inspection of sample images is the most intuitive approach. It is usually possible to quickly identify strengths and weaknesses, provided there is enough diversity in the images. However, it is also helpful to consider unbiased and objective metrics, such as those described in section \ref{sec:sota-image-comparison}.

\section{Results}
\subsection{Synthetic datasets}
\begin{figure}[htbp]
\centering\includegraphics[width=0.99\linewidth]{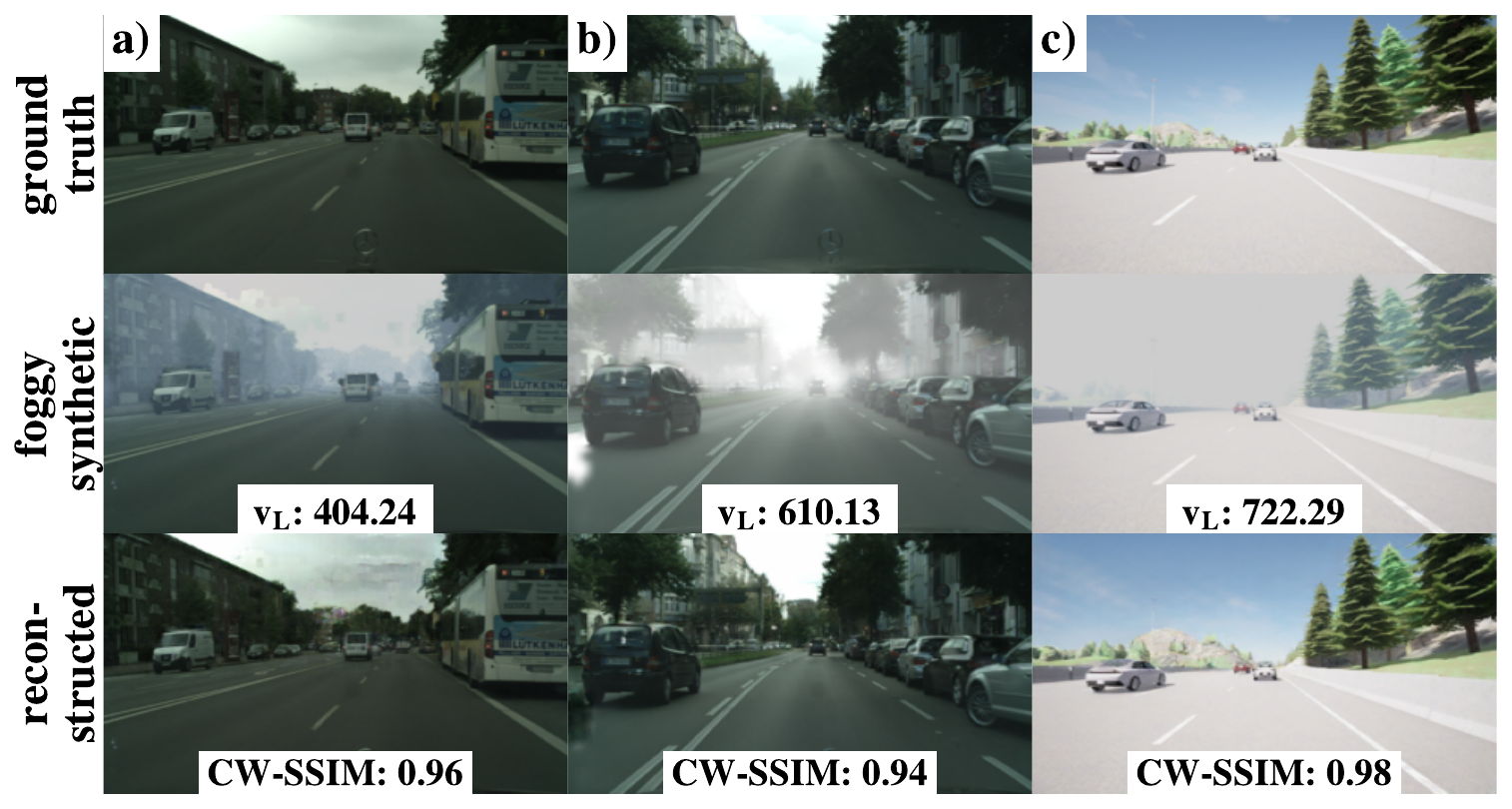}
\caption{Example results for the synthetic datasets. a): Cityscapes dataset from Uni. Tübingen \cite{Cordts2016, Bernuth2019}, b): Foggy Cityscapes dataset \cite{Cordts2016, Sakaridis2018}, c) Foggy CARLA dataset from Uni. Tübingen \cite{Dosovitskiy2017, Bernuth2019}}
\label{fig:synthetic-results}
\end{figure}

Figure \ref{fig:synthetic-results} shows the results of applying the \verb|pix2pix| model to the synthetic fog datasets described in section \ref{sota:fog-datasets}. The model exhibits very good performance on these datasets. This is expected, since the fog is computer-generated and computational defogging is relatively easy. The good performance is also evident in the CW-SSIM scores, which are close to perfect. The score for the CARLA dataset is slightly higher than that for the real-life based datasets, since a perfect depth map was available for fog generation.

\subsection{\textsc{Stereofog} dataset}
Figure S8 in Supplement 1 shows example evaluation results for the nine hyperparameter tests conducted, along with each model's metrics averaged across the entire testing set. Table S2 in Supplement 1 lists the best-performing value for each hyperparameter.

\begin{figure}[htbp]
\centering\includegraphics[width=0.99\linewidth]{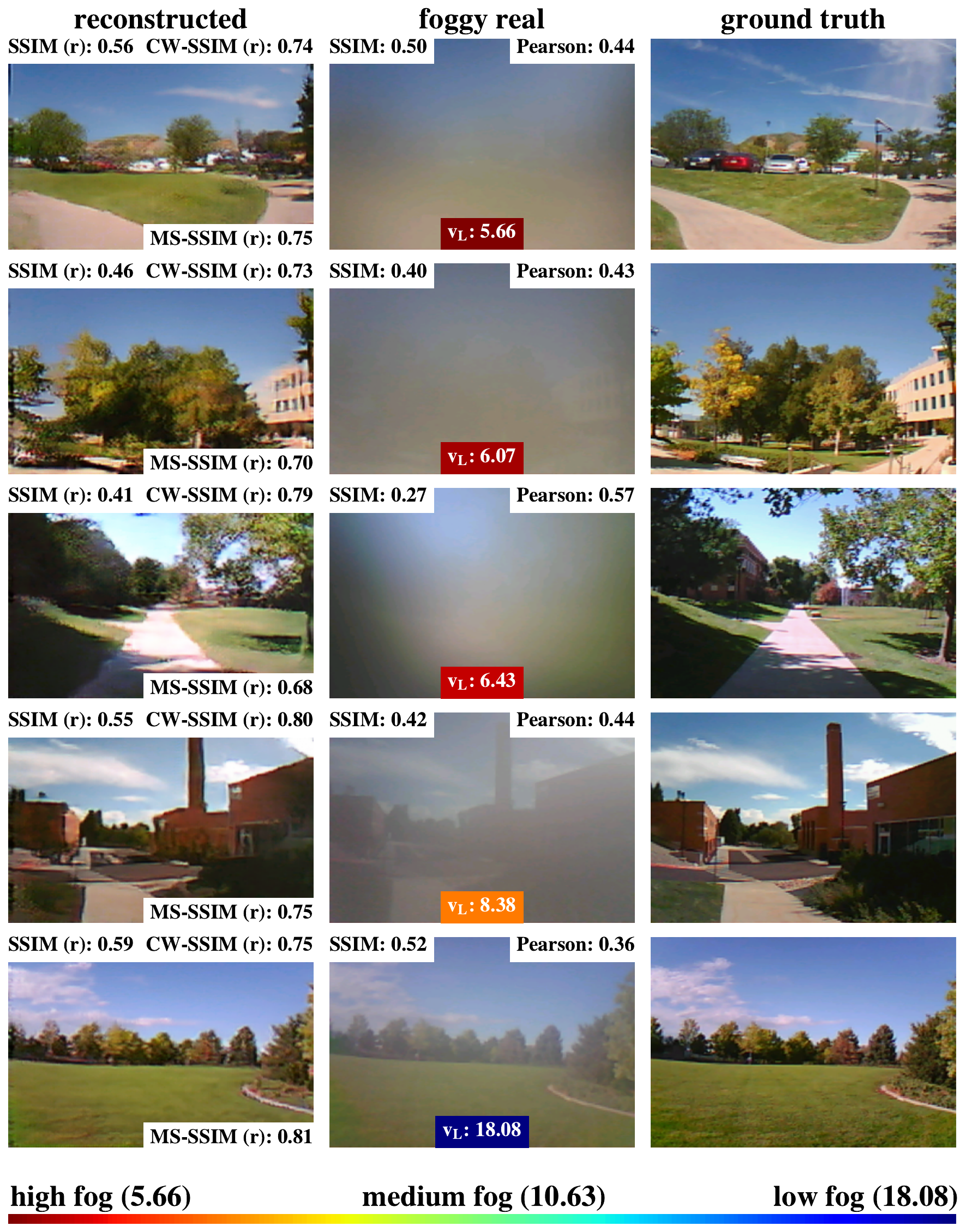}
\caption{Results from the \textsc{Stereofog} dataset with the optimum set of hyperparameters.}
\label{fig:stereofog-results}
\end{figure}

Exemplary defogged images using the best-performing model on the \textsc{Stereofog} dataset are shown in figure \ref{fig:stereofog-results}. The model is able to produce plausible reconstructions even when the input images have dense fog, such as in the top row. As expected, reconstructions from images with lighter fog are better. This best-performing model achieves the following average metrics:
\begin{gather}
    Pearson = 0.4;\
    MSE = 31.6;\
    NCC = 0.84;\ \nonumber\\
    SSIM = 0.41;\
    CW\mathrm{-}SSIM = 0.76;\
    MS\mathrm{-}SSIM = 0.7 \nonumber
\end{gather}

\subsection{Quality of reconstructions}
\begin{figure}[htbp]
\centering\includegraphics[width=\linewidth]{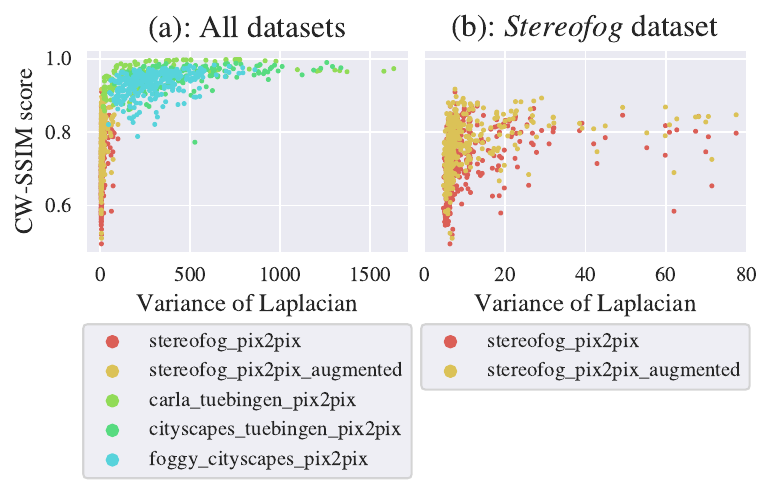}
\caption{Performance of the model against the fog density (measured as the variance of the Laplacian, $\text{v}_\text{L}$). Denser fog leads to lower $\text{v}_\text{L}$. (a) All datasets. (b) \textsc{Stereofog} dataset.}
\label{fig:reconstruction-quality}
\end{figure}

We investigated the effect of the fog density on the performance of our computational defogger, and the results are summarized in figure \ref{fig:reconstruction-quality}. As mentioned before, the variance of the Laplacian ($\text{v}_\text{L}$) is a measure of the fog density, which is labeled in each image. We quantified the performance of our defogger via the CW-SSIM (see section \ref{sec:sota-image-comparison}), which clearly drops as the fog density increases ($\text{v}_\text{L}$ decreases) as indicated in Figs. \ref{fig:reconstruction-quality}a and b for all datasets, and for the \textsc{Stereofog} dataset, respectively. We note that the \textsc{Stereofog} dataset exhibits higher fog densities (lower $\text{v}_\text{L}$), which leads to worse performance. When the fog density is low, the CW-SSIM score is high, $0.95$ for synthetic data and $0.8$ for the real data (\textsc{Stereofog}). But at higher fog densities, the scores can rapidly drop, to as low as $0.5$ in the case of the \textsc{Stereofog} dataset.

\section{Discussion}
In this paper, we introduce the \textsc{Stereofog} dataset comprised of foggy-clear image-pairs of real images that can be used to train machine-learning algorithms for computational defogging. Here, we specifically perform image-to-image translation via the \verb|pix2pix| framework. Although our results summarized in figure \ref{fig:stereofog-results} show significant promise, practical application of computational defogging requires further development as briefly outlined here.

\paragraph{Overexposure in dataset}
When creating the \textsc{Stereofog} dataset, we took care to avoid over-exposure due to glare. This was primarily due to the limited dynamic range of the chosen \emph{OpenMV H7} cameras. Further work with high-dynamic-range (HDR) imaging could improve the performance and applicability of our approach. 

\paragraph{Dataset diversity}
A second limitation of our work is the lack of diversity in the \textsc{Stereofog} dataset. Most images were composed of sunny scenes, since they were mostly collected during daytime in June to August, which are the clearest months in Salt Lake City \cite{CedarLakeVenturesInc.}. Enhancing the weather diversity of this dataset is expected to improve the model performance. Secondly, most of the images were collected on our University campus, which introduces bias. Both limitations mean that a model trained on this dataset will not perform well on images that depict weather conditions, locations, or features that are underrepresented or not represented at all in our dataset.

\paragraph{Dataset size}
The dataset is currently comprised of $10,067$ images. Even with augmentation, this is still relatively small in comparison to others such as the edges2handbags ($137,000$ images)  \cite{Zhu2018} and edges2shoes ($50,000$ images) \cite{Yu2014}. Increasing the number of images is expected to improve performance. 

\begin{figure}[htbp]
\centering\includegraphics[width=0.99\linewidth]{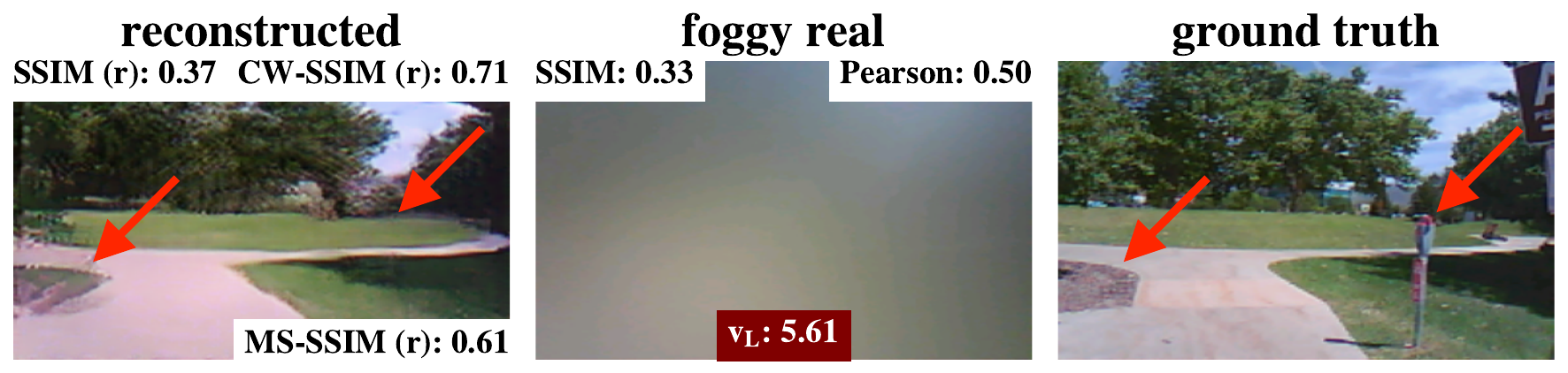}
\caption{Example of model bias in the final \textsc{Stereofog} model}
\label{fig:bias-example}
\end{figure}

\paragraph{Model bias}
The issue of model bias is closely related to the diversity and size of the dataset. Figure \ref{fig:bias-example} illustrates an example of model bias: The post that is clearly distinguishable in the ground truth image entirely disappears in the reconstructed one This can be explained by the fact that images with this type of pole are uncommon in our dataset, and the model is therefore biased to not recognize it. This is also the case for the brown area in the bottom left of the picture, which is reconstructed as partially green, since green lawn is more common in the dataset. It is important to note that the fog in the image also obscures the details of the pole and the brown patch, further impeding their accurate reconstruction. This issue can be improved through more diverse and larger dataset, and a better ML model.

\paragraph{Further hyperparameter tuning}
Within this work, the performances of the different hyperparameters were evaluated separately, without observing interdependencies between them. This was omitted due to the large amount of training necessary, but could be subject to further research.

\paragraph{Other algorithms}
A rigorous comparison of the performance of other types of algorithms would be worthwhile. Promising alternatives include the Restormer architecure \cite{Zamir2022} or the Lensless Imaging Transformer \cite{Pan2022}.

\paragraph{Image recognition tasks}
For many practical applications, it is useful to perform object identification and classification tasks on the defogged images. We show a preliminary result in this direction in Fig. S12 of the supplement using the \verb|Pixellib| Python library and a \emph{PointRend} model \cite{Olafenwa2020}.

\paragraph{Confidence quantification}
Another practical consideration is the quantification of the model's confidence in the defogged image. For example, this could be used in autonomous driving to detect whether the defogged image from the model is well-suited for a particular task, e.g., because the fog is too thick, and safely disengage the autonomous driving functions.

\section*{Acknowledgments}
This work was supported by a fellowship of the German Academic Exchange Service (DAAD). Furthermore, we would like to thank the researchers Georg Volk and J\"org Gamerdinger from the Chair of Embedded Systems at the University of T\"ubingen, Germany for guidance and use of their synthetic datasets. Additionally, we would like to thank the researchers Rich Baird, Al Ingold and Apratim Majumder for helpful guidance and discussion. Finally, we are also very thankful to the staff of the Maker Space at the University of Utah, whose facilities were used extensively throughout the project. The support and resources from the Center for High Performance Computing at the University of Utah are gratefully acknowledged.

\bibliographystyle{IEEEtran}
\bibliography{stereofog-paper}

\begin{IEEEbiography}[{\includegraphics[width=1in,height=1.25in,clip,keepaspectratio]{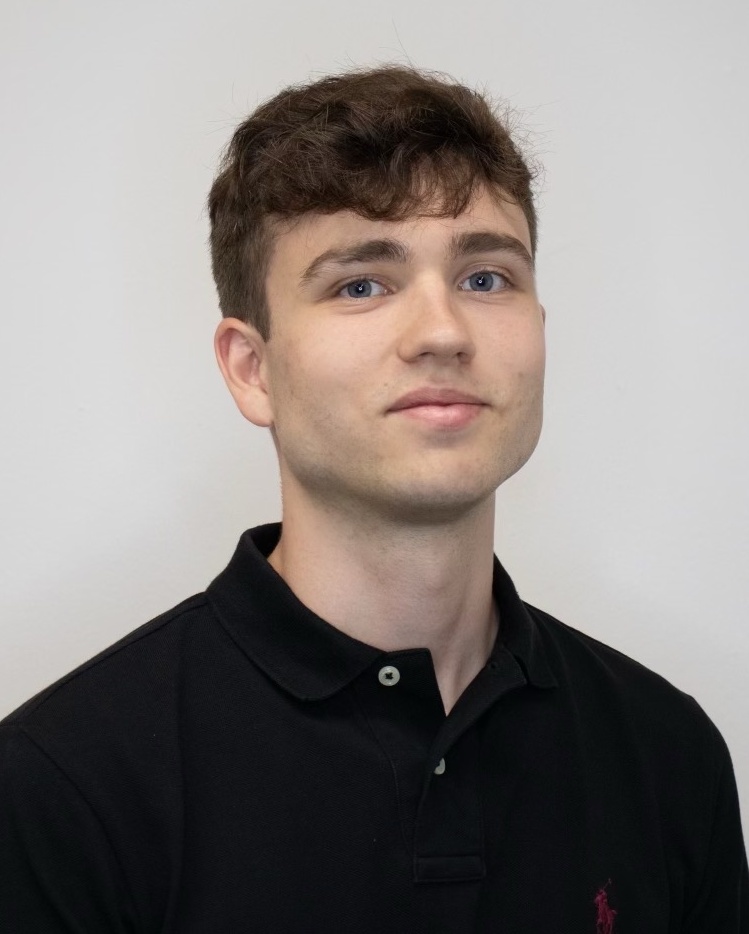}}]{Anton Pollak}
is currently studying mechanical engineering at Technical University of Berlin. As part of his degree, he spent a semester abroad at the University of Melbourne, Australia. He is working as a student assistant at the department Methods of Product Development and Mechatronics at the Technical University of Berlin, in the framework of the research projects MARBLE and zeroCUTS II. His research interests are robotics, machine learning as well as life cycle analyses in product development. He was supported by a DAAD RISE worldwide fellowship for undergraduate research at the University of Utah.
\end{IEEEbiography}

\begin{IEEEbiography}[{\includegraphics[width=1in,height=1.25in,clip,keepaspectratio]{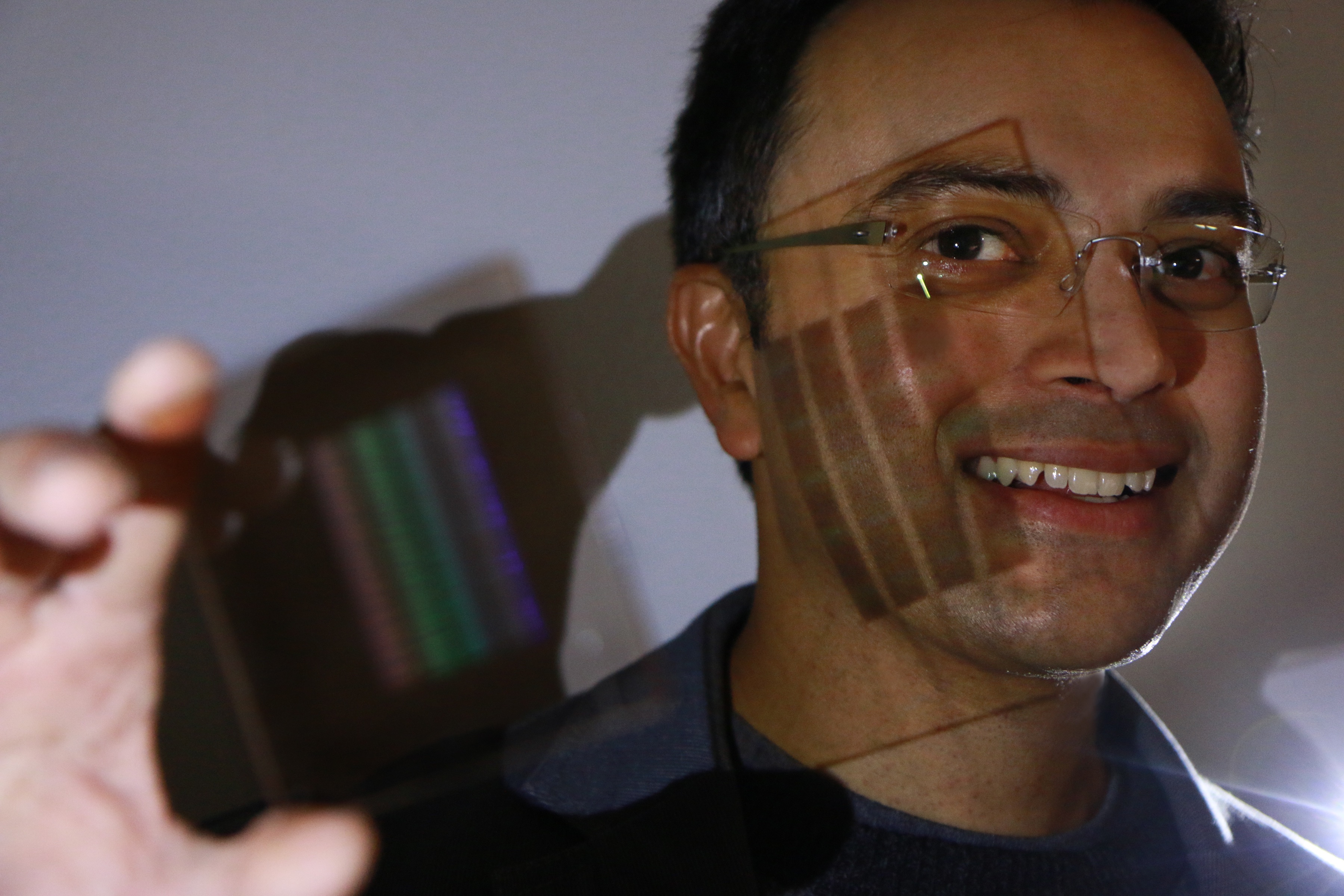}}]{Rajesh Menon}
combines his expertise in nanofabrication, computation and optical engineering to impact myriad fields including super-resolution lithography, metamaterials, broadband diffractive optics, integrated photonics, photovoltaics and computational optics. He is a Fellow of the Optical Society of America, and a Fellow of the SPIE, and a Senior Member of the IEEE. Among his other honors are a NASA Early Stage Innovations Award, NSF CAREER Award and the International Commission for Optics Prize.
\end{IEEEbiography}

\vfill

\end{document}